%% file: ms.tex
\documentclass[sigconf]{acmart}

\renewcommand\footnotetextcopyrightpermission[1]{} 
\usepackage{paper_preamble}
\usepackage{booktabs} 

\begin{document}

\title[Impact of the Cutoff Time on the Performance of Algorithm Configurators]{On the Impact of the Cutoff Time on the Performance of Algorithm Configurators}

\author{George T. Hall}
\affiliation{%
  \institution{Department of Computer Science}
  \city{University of Sheffield, Sheffield, UK}
}

\author{Pietro S. Oliveto}
\affiliation{%
  \institution{Department of Computer Science}
  \city{University of Sheffield, Sheffield, UK}
}

\author{Dirk Sudholt}
\affiliation{%
  \institution{Department of Computer Science}
  \city{University of Sheffield, Sheffield, UK}
}

\input{sections/abstract.tex}

%
\keywords{Parameter tuning, Algorithm configurators, Runtime analysis}
 

\maketitle

\subfile{sections/intro}

\subfile{sections/preliminaries}

\section{{\scshape ParamRLS} for RLS$_{\text{\boldmath{$k$}}}$ and {\scshape Ridge*}}
\label{sec:tuning_for_ridge}
In this section we will prove that ParamRLS-F identifies the optimal parameter $k=1$ for RLS$_k$ and \rdgst for any cutoff time.
If the cutoff time is large enough i.e., $\kappa=\omega(n)$, then even just one run per configuration evaluation suffices.
For smaller cutoff times,  ParamRLS-F requires more runs per configuration evaluation to identify that RLS$_1$ is better than any other RLS$_k$ for $k>1$.
We will show this for the extreme case $\kappa=1$ for which $n^{3/2}$ runs per evaluation suffice for ParamRLS-F to identify the correct parameter w.o.p.
On the other hand, ParamRLS-T will return a random configuration for any $\kappa<n^2/2$.
The range of parameter values goes up to $\phi = \sqrt{n}$; larger values of $k$ degrade to random search.

\subfile{sections/rlsk_ridge_analysis}

\subfile{sections/ridge_relative_performance}
\subfile{sections/ridge_tuner_analysis}

\section{{\scshape ParamRLS} for RLS$_{\text{\boldmath{$k$}}}$ and {\scshape OneMax}}
\label{sec:tuning_for_om}

\subfile{sections/rlsk_om_analysis}
\subfile{sections/om_long_races}

\subfile{sections/om_short_races}

\subfile{sections/conclusion}

\paragraph{Acknowledgements} This work was supported by the EPSRC under grant EP/M004252/1.

\bibliographystyle{plain}
\bibliography{bibliography}

\subfile{sections/appendix}

\end{document}

%% file: sections/abstract.tex
\begin{abstract}

    Algorithm configurators are automated methods to optimise the parameters of
    an algorithm for a class of problems. We evaluate the performance of a
    simple random local search configurator (ParamRLS) for tuning the
    neighbourhood size $k$ of the RLS$_k$ algorithm. We measure performance as
    the expected number of configuration evaluations required to identify the
    optimal value for the parameter. We analyse the impact of the cutoff time
    $\kappa$ (the time spent evaluating a configuration for a problem instance) on
    the expected number of configuration evaluations required to find the
    optimal parameter value, where we compare configurations using either best
    found fitness values (ParamRLS-F) or optimisation times (ParamRLS-T). We
    consider tuning RLS$_k$ for a variant of the {\scshape Ridge} function
    class ({\scshape Ridge*}), where the performance of each parameter value
    does not change during the run, and for the {\scshape OneMax} function
    class, where longer runs favour smaller $k$. We rigorously prove that
    ParamRLS-F efficiently tunes RLS$_k$ for {\scshape Ridge*} for any  $\kappa$ while
    ParamRLS-T requires at least quadratic $\kappa$. For {\scshape OneMax}
    ParamRLS-F identifies $k=1$ as optimal with linear $\kappa$ while
    ParamRLS-T requires a $\kappa$ of at least $\Omega(n\log n)$. For smaller
    $\kappa$ ParamRLS-F identifies that $k>1$ performs better while ParamRLS-T
    returns $k$ chosen uniformly at random.

\end{abstract}

%% file: sections/intro.tex
\section{Introduction}
General purpose heuristics, such as evolutionary algorithms, have the advantage that they can generate high quality solutions to optimisation
problems without requiring much knowledge about the problem at hand. All that is required to apply a general purpose heuristic is a suitable representation for candidate solutions
and a measure (the fitness function) that allows us to compare the quality of different solutions against each other.
However, it is well understood that different design choices and different settings of their numerous parameters (e.g., mutation rate, crossover rate, selective pressure and population size for generational genetic algorithms (GAs)) may considerably affect their performance and in turn the quality of the identified solutions.
In particular, the capability of heuristics to identify high quality solutions in a short time depends crucially on the use of suitable parameter settings~\cite{paper:EibenParameterControl}.

Traditionally the design and parameter tuning of the algorithm for the problem at hand has mainly been done {\it manually}.
Typically, the developer chooses some algorithmic designs and values for the associated parameters and executes them on instances of the problem.
Refinements are then made according to how well each algorithm/parameter configuration has performed.

However, such a procedure (or a similar one) is a time-consuming and error-prone process.
From a scientific research point of view, it is also biased by personal experience hence difficult to reproduce.
Consequently it has become increasingly
common to use automated and principled methodologies for algorithm development.
In the literature, researchers have typically referred to the automated optimisation of algorithm performance as {\it automated parameter tuning}  and {\it automated algorithm configuration} \cite{chap:stutzle_lopez_ibanez}.
Recently more ambitious methodologies have emerged such as {\it automated construction of heuristic algorithms}~\cite{SATenstein,paper:Fukunaga2008}  {\it automated algorithm generation}~\cite{paper:paramILS} and {\it hyper-heuristics}~\cite{BurkeEtAl2013}.

Although automating the algorithmic design has gained significant momentum in recent years, the idea has been around for over thirty years.
In 1986 Grefenstette used a GA to optimise the parameters of another GA \cite{paper:meta_GA_param_tuning}. Since then several other heuristic methodologies have been employed
to optimise algorithmic parameters including hill-climbing \cite{paper:analysis_learning_plan_search_problem}, 
beam search \cite{paper:integrating_heuristics_constraint_satisfaction_probs}, iterated local search (ParamILS)~\cite{paper:paramILS}, gender-based GAs \cite{paper:gender_based_GA_param_tuner} and more traditional GAs (EVOCA) \cite{paper:new_algo_reduce_metaheuristic_effort}.
Recently more sophisticated methodologies have appeared based on {\it racing} \cite{paper:racing_introduced} approaches for comparing several configurations in parallel and integrating statistical testing methods \cite{paper:f_race_introduced}. These include the popular {\it irace} configurator \cite{paper:irace}. Also surrogate models have been introduced to predict the computational cost of testing specific configurations in order to avoid poor choices. Popular examples of surrogate-based configurators are sequential parameter optimisation (SPOT) \cite{paper:SPO, paper:SPOT} and the sequential model-based algorithm configuration (SMAC) \cite{paper:ROAR_and_SMAC}.

While varying in several algorithmic details, all algorithm configurators generally aim to evolve better and better parameter values by evaluating the performance of candidate configurations on a
training set of instances and using some perturbation mechanism (e.g., iterated local search in ParamILS or updating the sampling distributions in irace) to generate new ones based on the better performing ones in the previous generation. The overall aim is that the ultimately identified parameter values perform well ({\it generalise}) on unseen instances of the tackled problem.
Many of the mentioned algorithm configurators have gained widespread usage since they have often identified better parameter values compared to carefully chosen default configurations \cite{paper:irace, paper:paramILS, paper:SPO, paper:SPOT, paper:ROAR_and_SMAC}.

Despite their popularity, there is a lack of theoretical
understanding of such configurators.
For instance, it is unclear how good the identified parameters are compared to optimal ones for a given target algorithm and optimisation problem.
In particular, if optimal parameter values may be identified by a given configurator, no indications are available regarding how large the total tuning budget should be for the task.
Similarly, it is unclear how long should each configuration be run for (i.e., {\it cutoff time}) when evaluating its performance on a training set instance.

In this paper, we take a first step towards
establishing a theoretical grounding of algorithm configurators. 
Similarly to the time complexity analysis of other fields~\cite{AugerDoerr} we begin by analysing simplified algorithms and problems with the aim of building up a set of mathematical techniques for future analyses
of more sophisticated systems and to shed light on for which classes of problems more sophistication is required for good performance.

We consider a simple hillclimbing tuner, which we call ParamRLS because
it is a simplified version of the popular ParamILS tuner.
The tuner mutates the value of one of its parameters chosen uniformly at random to create an offspring configuration which will be accepted if it performs at least as well as its parent on the training set.
Regarding configuration performance evaluations, we consider two versions of ParamRLS.
One, ParamRLS-T, compares the average runtimes required by the different configurations to identify the optimal solution of the target instances.
If the instance is not solved by a configuration, then the cutoff time is returned multiplied by a penalty factor called {\it penalisation constant}. This performance measure originates in the SAT community, where it is called \emph{penalised average runtime (PAR)} \cite{SATenstein}. The other version, ParamRLS-F, compares the number of times that solutions of better fitness are identified within the cutoff time by the different configurations and breaks ties by preferring the configuration that took less time to identify them.
We analyse time-based comparisons because they are typically used in ParamILS, and are also available in SMAC and irace. We compare them with the latter strategy.

While the tuner is very simple, the mathematical methods developed for its analysis are quite sophisticated and  can be built upon for the analysis of more complicated algorithm configurators since the performance comparison of (at least) two parameter configurations is at the heart of virtually any parameter tuner.
To the best of our knowledge, this is the first time that a rigorous time complexity analysis
of algorithm configurators has been performed. The only related theoretical work regards
the performance analysis of (online) parameter control of randomised search heuristics
during the function optimisation phase~\cite{AlanaziLehre2014,DoerrEtAl2016B,DLOW2018,LehreOzcan2013,LissovoiEtAl2019,QianEtAl2016,LOWGecco2017,LOWArxiv2018}.

We will analyse the number of iterations required  by ParamRLS to identify optimal parameter values with overwhelming probability (\wop)\footnote{We say that a probability is overwhelming if it is at least $1-2^{-\Omega(n^\varepsilon)}$ for some constant~${\varepsilon > 0}$. We frequently use that by a union bound, any polynomial number of events that all occur \wop occur together with overwhelming probability.} for the randomised local
search (RLS$_k$) algorithm, where $k$, the only parameter, is the local search neighbourhood size (i.e., $k$ bits are flipped without replacement in
each iteration). Our aim is to characterise the impact of the cutoff time on the performance of the tuner.
We will perform the analysis for two well-known black-box benchmark function classes: a
modified version of {\scshape Ridge} (called \rdgst[)] and {\scshape OneMax}\footnote{The  {\scshape OneMax} function class consists of $2^n$ functions over $\{0,1\}^n$ each with a different global optimum and for each function the fitness decreases with the Hamming distance to the optimum.}~\cite{DrosteJansenWegener2002}.
%
%
Since for both function classes, a given parameter configuration will have the same performance for all instances,
these classes allow us to avoid the problems of deciding how many instances should be used in the training set (i.e., one instance suffices) and of evaluating the generalisation capabilities
of the evolved parameters (i.e., the performance will be the same for all instances). Hence, we can concentrate on the impact of the cutoff time in isolation.

The two function classes have different characteristics.
For \rdgst[,] each parameter value has the same improvement probability independent of the position of the candidate solution in the search space.
For {\scshape OneMax}, it is better to flip fewer bits the
closer the candidate solution is to the optimum.
This implies that for \rdgst the optimal parameter value is the same independent of how long the algorithm is run for i.e., $k=1$ will have better performance even for very small cutoff times
as long as a sufficient number of comparisons between different configurations are performed. For {\scshape OneMax},
 short runs of RLS$_k$ with larger values of $k$ finds better solutions, whereas for longer runs
smaller values of $k$ perform better.  \par

Our analysis shows that ParamRLS-F can efficiently identify that $k=1$ is the optimal parameter value for \rdgst independent of the cutoff time as long as the performance for each parameter configuration is evaluated a sufficient number of times. For {\scshape OneMax}, instead, ParamRLS-F identifies that $k=1$ is the optimal parameter for any cutoff time greater than $4n$. If the cutoff time is considerably smaller, then ParamRLS-F will identify that the optimal value is $k>1$.
On the other hand, ParamRLS-T returns a parameter value chosen uniformly at random for any function containing up to an exponential number of optima if the cutoff time is smaller than $(n \ln n)/2$. We show that for \rdgst the cutoff time for ParamRLS-F has to be at least quadratic in the problem size.




This paper is split into three sections. In Section~\ref{sec:prelims}, we
describe the algorithm configuration problem, the algorithms and the
function classes considered in this paper. We analyse ParamRLS tuning RLS$_k$ for
{\scshape Ridge*} and {\scshape OneMax} in Sections~\ref{sec:tuning_for_ridge}~and~\ref{sec:tuning_for_om}, respectively. Some proofs are omitted from the main part of the paper due to space restrictions. The omitted proofs from the main part can be found in the appendix.


%% file: sections/preliminaries.tex
\section{Preliminaries}

\label{sec:prelims}

\subsection{The Algorithm Configuration Problem} 

Informally, given an algorithm $\mathcal{A}$, its set of parameters $\theta_{\mathcal{A}} = \{P_1, \dots, P_{N_P}\}$
and an optimisation problem $\mathcal{P}$, the \emph{algorithm configuration problem} is that of identifying the set of parameter values $\theta^*$ 
for which $\mathcal{A}$ achieves best performance on $\mathcal{P}$.
We call the algorithm solving the configuration problem the \emph{configurator} and the algorithm to
be tuned ($\mathcal{A}$)  the \emph{target algorithm}\footnote{Note that throughout the paper we use the terms \emph{configurator}
and \emph{tuner} interchangeably.}.
%

More formally, we use $\Theta$ to denote the
\emph{parameter configuration space}  of $\mathcal{A}$ (i.e., the search space of all feasible parameter configurations)
and we denote a specific configuration by $\theta \in \Theta$.
The performance of different configurations for the problem $\mathcal{P}$ is evaluated on a \emph{training set}  of instances $\Pi$ 
which should be representative of the problem. 
Finally, let $cost$ be a measure of the performance of running $\mathcal{A}(\theta)$ over the training set $\Pi$.
Then the algorithm
configuration problem is that of finding
\[ \theta^* \in \arg \min_{\theta \in \Theta} cost(\theta) \]

The $cost$ function estimates the performance of algorithm $\mathcal{A}$ on a training set of problem instances $\Pi$.
To do so the following decisions need to be made:
\begin{itemize}
\item Which instances (and how many) should be used in the training set $\Pi$;
\item  \emph{Cutoff time} $\kappa$: the amount of time that the algorithm $\mathcal{A}$ is run on each instance $\pi_i \in \Pi$;
\item   Runs $r$: the number of times the evaluation (of duration $\kappa$) should be repeated for each instance $\pi_i \in \Pi$;
\item   $Metric$: the quantity that is measured to evaluate how well  $\mathcal{A}(\theta)$ performs on each $\pi_i \in \Pi$;
\item How to aggregate the measure of performance over all instances.
\end{itemize}

Since for the two instance classes considered in this paper (see Section \ref{sec:problems}) one random instance suffices for perfect generalisation\footnote{\emph{Perfect generalisation} means that the algorithm configuration will work equally well on problem instances that are not in the training set.}, we do not need to worry about the choice of the training set nor how to aggregate performances over it.
We will consider two different \emph{metrics}:
\begin{enumerate}
\item The time required  for $\mathcal{A}(\theta)$ to find the optimal solution of an instance $\pi_i$. If the optimum is not found before the cutoff time $\kappa$, then $p  \cdot \kappa$ is taken
    as the time to reach the optimum, where $p$ is a penalty constant. This metric is commonly used in ParamILS~\cite{paper:paramILS}. 
\item The fitness of the best solution found within the cutoff time.
\end{enumerate}

Let $T$ be the number of tested configurations before the optimal configuration $\theta^*$ is identified. We call this the number of evaluated configurations, or the number of evaluations.
Then the total tuning time will be $\mathcal{B} = T \cdot |\Pi| \cdot \kappa \cdot r$. \par
Our aim in this paper is to estimate, for each metric, how the cutoff time $\kappa$ and the number of runs $r$ impact the number of evaluated configurations $T$ and the total
tuning time $\mathcal{B}$ for a simple configurator called ParamRLS.



\subsection{The Configurator: ParamRLS}

We design our simple configurator following the framework laid out 
for ParamILS \cite{paper:paramILS}:
\begin{enumerate}
\item Initialise the configurator with some initial configuration $\theta$;
\item mutate $\theta$ by modifying a single parameter and accept the new configuration $\theta'$ if it results in improved performance;
\item repeat Step 2 until no single parameter change yields an improvement.
\end{enumerate}

Essentially we follow the above scheme where we initialise the configurator choosing a configuration uniformly at random from $\Theta$ and we change the acceptance criterion to accept a new configuration if it performs at least as well as its parent. Note that we occasionally refer to the current value of $\theta$ in Algorithm~\ref{algo:pRLS} as the \emph{active parameter}. Concerning Step 2, ParamILS applies an Iterated Local Search procedure.

We instead consider the following two more simple random local search operators and, thus, call the algorithm ParamRLS:
\begin{itemize}
    \item{$\pm 1$: the chosen parameter value is increased or decreased by 1 uniformly at random;}
    \item{$\pm \{1,2\}$: the chosen parameter value is increased or decreased by 1 or by 2 uniformly at random.}
\end{itemize}
The first operator has previously been analysed for the optimisation of functions defined over search spaces with larger alphabets than those that can be represented using bitstrings~\cite{paper:DoerrDoerrKoetzing16}. The second one slightly enlarges the neighbourhood size.
For both operators we use the interval-metric such that any mutation that oversteps a boundary is considered infeasible.
The resulting configurator is described in Algorithm~\ref{algo:pRLS}.
The termination condition may be either a predetermined number of iterations without a change  in configuration (i.e., the solution is likely a local or global optimum) or a fixed number of iterations.
In this paper we calculate the number of iterations until the configurator identifies the optimal configuration and will not leave it with overwhelming probability, hence we also provide bounds on the termination criterion.

If the configurator uses the fitness-based metric for performance evaluation described in the previous section, then we will call the algorithm ParamRLS-F while if it uses the time-based metric, then we will refer to it as ParamRLS-T. The two evaluation procedures are described respectively in Algorithm~\ref{algo:evalF} and in Algorithm~\ref{algo:evalT}. In Algorithm~\ref{algo:evalT}, we denote the capped optimisation time for $\mathcal{A}(\theta)$ on $\pi_i$ with cutoff time $\kappa$ and penalty constant $p$ as \texttt{CappedOptTime}$(\mathcal{A}(\theta, \pi_i), \kappa, p)$.

\begin{algorithm}[t]

    \begin{algorithmic}[1]
        \STATE{$\theta \gets $initial parameter value chosen uniformly at random}
        \WHILE{termination condition not satisfied}
            \STATE{$\theta' \gets \text{\texttt{mutate($\theta$)}}$} 
                \STATE{$\theta \gets \text{\texttt{eval}($\mathcal{A},\theta,\theta',\kappa,r$)}$}
        \ENDWHILE
        \RETURN $\theta$
    \end{algorithmic}

    \caption{ParamRLS ($\mathcal{A},\Theta,\Pi,\kappa,r$)}
    \label{algo:pRLS}

\end{algorithm}

%
%
%

\begin{algorithm}[t]

    \begin{algorithmic}[1]
        \STATE $Wins \gets 0$; $Wins' \gets 0$ \COMMENT{count number of wins for $\theta$ and $\theta'$}
        \STATE $R \gets 0$
        \WHILE {$R < r$}
            \STATE $ImprovementTime \gets 0$
            \STATE $ImprovementTime' \gets 0$
            \STATE $Fitness \gets \mathcal{A}(\theta, \pi_i)$ fitness after $\kappa$ iterations; 
            \STATE $Fitness' \gets \mathcal{A}(\theta', \pi_i)$ fitness after $\kappa$ iterations; 
            \STATE $ImprovementTime \gets $time of last impr of $\mathcal{A}(\theta, \pi_i)$
            \STATE $ImprovementTime' \gets $time of last impr of $\mathcal{A}(\theta', \pi_i)$
            \IF{$Fitness > Fitness'$}
                \STATE $Wins \gets Wins + 1$
            \ELSIF{$Fitness' > Fitness$}
                \STATE $Wins' \gets Wins' + 1$
            \ELSE
                \IF{$ImprovementTime < ImprovementTime'$}
                    \STATE $Wins \gets Wins + 1$
                \ELSIF{$ImprovementTime' < ImprovementTime$}
                    \STATE $Wins' \gets Wins' + 1$
                \ENDIF
            \ENDIF
            \STATE  {$R \gets R+1$}
        \ENDWHILE
        \STATE \algorithmicif\ $Wins > Wins'$ \algorithmicthen\ \algorithmicreturn\ $\theta$
        \STATE \algorithmicelsif\ $Wins' > Wins$ \algorithmicthen\ \algorithmicreturn\ $\theta'$
        \STATE \algorithmicelse\ \algorithmicreturn\ a uniform choice of $\theta$ or $\theta'$
    \end{algorithmic}

    \caption{The \texttt{eval-F}($\mathcal{A},\theta,\theta',\pi_i,\kappa,r$) subroutine in ParamRLS-F}
    \label{algo:evalF}

\end{algorithm}

\begin{algorithm}[t]

    \begin{algorithmic}[1]
        \STATE $Time \gets 0$; $Time' \gets 0$ \COMMENT{count optimisation times for $\mathcal{A}(\theta, \pi_i)$ and $\mathcal{A}(\theta', \pi_i)$}
        \STATE $R \gets 0$
         \WHILE {$R < r$}
            \STATE $Time \gets Time + \text{\texttt{CappedOptTime}}(\mathcal{A}(\theta, \pi_i), \kappa, p)$\!\!\!\!\!\!\!\!\!\!\!\!\!\!\!\!\!\!\!\!
            \STATE $Time' \gets Time' + \text{\texttt{CappedOptTime}}(\mathcal{A}(\theta', \pi_i), \kappa, p)$\!\!\!\!\!\!\!\!\!\!\!\!\!\!\!\!\!\!\!\!
            \STATE  {$R \gets R+1$}
        \ENDWHILE
        \STATE \algorithmicif\ $Time < Time'$\ \algorithmicthen\ \algorithmicreturn\ $\theta$
        \STATE \algorithmicelsif\ $Time' < Time$\ \algorithmicthen\ \algorithmicreturn\ $\theta'$
        \STATE \algorithmicelse\ \algorithmicreturn\ a uniform choice of $\theta$ or $\theta'$
    \end{algorithmic}

    \caption{The \texttt{eval-T}($\mathcal{A},\theta,\theta',\pi_i, \kappa,r$) subroutine in ParamRLS-T}
    \label{algo:evalT}

\end{algorithm}

\subsection{The Target Algorithm: RLS$_{\text{\boldmath{$k$}}}$}
In this paper we will evaluate the ParamRLS configurator for tuning the RLS$_k$ algorithm
which has only one parameter $k$.
RLS$_k$ differs from conventional RLS in that the latter flips exactly one bit per
iteration whereas RLS$_k$ flips exactly $k$ bits per iteration, selected
without replacement. Our aim is to identify the time required by our simple
tuner to identify the best value for the parameter $k$. We provide the pseudocode for RLS$_k$ in
Algorithm~\ref{algo:RLSk}. We define the permitted values for $k$ as the range $1, \ldots, \phi$.

\begin{algorithm}[t]
    \begin{algorithmic}[1]

        \STATE{\textbf{initialise $x$}} \COMMENT{according to initialisation scheme} \label{line:init}
        \WHILE{termination criterion not met}
        \STATE{$x' \gets$ $x$ with $k$ distinct bits flipped}
        \STATE{\algorithmicif\ $f(x') \ge f(x)$ \algorithmicthen\ $x \gets x'$} \label{line:fxprime_ge_fx}
        \ENDWHILE

    \end{algorithmic}

    \caption{{\scshape RLS}$_{k}$ for the maximisation of a function $f$}
    \label{algo:RLSk}

\end{algorithm}

\subsection{The Function Classes {\scshape Ridge*} and {\scshape OneMax}} \label{sec:problems}
We will analyse the performance of ParamRLS for tuning RLS$_k$ for two optimisation problems with considerably different characteristics. One where the performance of each parameter configuration does not change throughout the search space and another where according to the cutoff times different configurations will perform better.

For the first problem we consider a modified version of the standard  {\scshape Ridge} benchmark problem~\cite{DrosteJansenWegener2002}. The conventional {\scshape Ridge} function
consists of a gradient of increasing fitness with the increase of the number of 0-bits in the bitstring that leads towards the $0^{n}$ bit string (i.e., \textsc{ZeroMax}).
From there a path of $n$ points, consisting of consecutive 1-bits followed only by 0-bits,
may be found that leads to the global optimum (i.e. the $1^{n}$ bit string).
To achieve the sought behaviour and at the same time simplify the analysis, we remove the \textsc{ZeroMax} part by assuming that the algorithm is initialised in the $0^{n}$ bit string. This technique
was used by Jansen and Zarges in order to simplify their early fixed budget analyses \cite{paper:fixed_budget_analysis}.
As a result any bit string not in the form $1^{i}0^{n-i}$ will
be rejected. We call our modified function {\scshape Ridge*}: 
\[
    \text{{\scshape Ridge*}}(x) =
        \begin{cases}
            i, \text{ if } x \text{ in form } 1^{i}0^{n-i} \\
            -1, \text{ otherwise}
        \end{cases}
\]
Since we are using RLS$_k$ to optimise {\scshape Ridge*}, it will not always be
possible to reach the optimum (i.e.\ $1^{n}$). The optimal value of {\scshape
Ridge*} which we are able to reach when using RLS$_k$ is in fact
$\floor{\frac{n}{k}}k$. In this work, we will consider reaching this value as
having optimised the function.

The black box optimisation version of {\scshape Ridge*} consists of $2^n$ functions. For each $a \in \{0,1\}^n$ the fitness of a solution $x$ for the corresponding function can be calculated using the following XOR
transformation: {\scshape Ridge}$^{*}_{a} (x) :=$ {\scshape Ridge}$^* (x_1 \oplus a_1 \dots x_n \oplus a_n)$~\cite{DrosteJansenWegener2002b}. For convenience of analysis we will use the {\scshape Ridge}$^{*}_{0^n}$ function displayed above where the path starts in the $0^{n}$ bit string and terminates in the $1^{n}$ bit string.
The best parameter value for RLS$_k$ for a random instance will naturally be optimal also for any other instance of the black box class.

The second optimisation problem we will consider is the well-studied {\scshape OneMax} benchmark function. Its black box class consists of $2^n$ functions each of which has a different bit string as global optimum and the fitness of each other bit string decreases with the Hamming distance to the optimum. We tune the parameter for only one instance since the  identified optimal parameter will naturally also be the best parameter for any of the other $2^n$ instances. In particular, we will use the instance: $\text{{\scshape OneMax}}(x) = \sum_{i=1}^{n} x_{i}$. 
%


\subsection{A General Result for ParamRLS-T}
In this section we show that for ParamRLS-T the cutoff time has to be at least superlinear in the instance size or it will not work.
We can show that, for any 
$\kappa \le (n \ln n)/2$ and any function with up to an exponential number of optima, ParamRLS-T with
overwhelming probability will return a parameter value chosen uniformly at random, for any
polynomial number of evaluations and runs per evaluation.
In Section \ref{sec:tuning_for_ridge} we will show that $\kappa$ has to be at least quadratic for ParamRLS-T to identify the optimal configuration of RLS$_k$ for {\scshape Ridge*}.

\begin{theorem}
    For RLS$_k$ on any function with up to $\exp(\sqrt{n}/\log^2 n)$ optima, ParamRLS-T
    with cutoff time $\kappa \le (n \ln n)/2$, local search operator $\pm 1$ or
    $\pm \{1,2\}$, and any polynomial number of evaluations $T$ and runs per
    evaluation $r$, will return a value for $k$ chosen uniformly at random, with overwhelming probability.
\end{theorem}

\begin{proof}
    Note that RLS$_k$ belongs to the class of unary unbiased black-box algorithms as defined in~\cite{Lehre2012}. Then~\cite[Theorem~20]{paper:parallel_black_box_complexity_tail_bounds} (applied with $\delta := 1/2$) tells us that all RLS$_k$ algorithms require at
    least $(n \ln n)/2$ iterations to reach the optimum, with probability $1 -
    \exp(-\Omega(\sqrt{n}/\log n))$.
    By the union bound, the probability that none of the $T \cdot r$
    total runs of RLS$_k$ reaches the optimum within $(n \ln n) / 2$ iterations
    is at least $1 - T \cdot r \cdot \exp(-\Omega(\sqrt{n}/\log n))$, which is again
    overwhelming for any polynomial choices of $T$ and $r$.
    This implies that the tuner has no information to guide the search
    process, and therefore accepts the new value of $k$ with probability 0.5.
    It is easy to show that the tuner returns a value for $k$ uniformly at
    random.
\end{proof}

%% file: sections/rlsk_ridge_analysis.tex

\subsection{Analysis of RLS$_{\text{\boldmath{$k$}}}$ on {\scshape Ridge*}}

\label{sec:analysis_rlsk_ridge}
In this section we analyse how the performance of RLS$_k$ for \rdgst changes with the parameter $k$.


\begin{lemma}
    For $k \le n/2$, the expected optimisation time of \rlsk on \rdgst is
    $\Floor{\frac{n}{k}} {n \choose k}$.
    \label{lem:exp_opt_time}
\end{lemma}

\begin{proof}
    During a single iteration, it is only possible to increase the fitness of
    an individual by exactly $k$ since we must flip exactly the first $k$
    zeroes in the bit string (any other combination of flips will mean that the
    string is no longer in the form $1^{i}0^{n-i}$ and will be rejected). We
    call an iteration in which we flip exactly the first $k$ zeroes in the bit
    string a \emph{leap}. There are ${n \choose k}$ possible ways in which we
    can flip $k$ bits and exactly one of these combinations
    flips the first $k$ zeroes. Therefore the probability of making a leap at
    any time $T$ is $1/{n \choose k}$. \par

    By the waiting time argument, we wait ${n \choose k}$ iterations in
    expectation to make a single leap. Since we need to make
    $\floor{\frac{n}{k}}$ leaps in order to reach the optimum, we wait
    $\floor{\frac{n}{k}} {n \choose k}$ iterations in expectation until we
    reach the optimum.
\end{proof}

\begin{corollary}
    A value of $k=1$ leads to the shortest expected optimisation time for \rlsk
    on \rdgst for any $k \le n/2$.
    \label{cor:k1_shortest_runtime}
\end{corollary}

The optimisation time is also highly concentrated around the expectation, with deviations by (say) a factor of 2 having an exponentially small probability. The following lemma follows directly from Chernoff bounds.

\begin{lemma}
    \label{lem:opt_lower_bound}
    With probability at least $1-\exp (-\Omega(n/k))$, RLS$_k$
    requires at least ${n \choose k} \lfloor n/k \rfloor / 2$ and at most $2{n \choose k} \lfloor n/k \rfloor$ iterations to
    optimise \rdgst[.]
\end{lemma}

%% file: sections/ridge_relative_performance.tex
\label{sec:relative_performance}

We can now consider the relative performance of RLS$_a$ and RLS$_b$ on
{\scshape Ridge*}, for some $a<b$. We first derive a general bound which can be
applied to any two random processes with probabilities of improving which stay
the same throughout the process. We derive a lower bound on the probability
that the process with the higher probability of improving is ahead at some time
$t$. We apply this to RLS$_a$ and RLS$_b$ for {\scshape Ridge*}.

\begin{lemma}
\label{lem:const_probs_diff_bound}
    Let $\mathcal{A}$ and $\mathcal{B}$ be two random processes which both take values from the non-negative real numbers, and both
    start with value 0. At each time step, $\mathcal{A}$ increases by some real number
    $\alpha \ge 0$ with probability $p_{a}$, and otherwise stays put. At each time
    step, $\mathcal{B}$ increases by some real number $\beta \ge 0$ with probability
    $p_{b}$, and otherwise stays put. Let $\Delta_t^a$ and $\Delta_t^b$ denote
    the total progress of $\mathcal{A}$ and $\mathcal{B}$ in $t$ steps,
    respectively. Let $q := p_a(1-p_b)+(1-p_a)p_b$, $q_a := p_a(1-p_b)/q$, and
    $q_b := p_b(1-p_a)/q$. Then, for all $0 \le p_b \le p_a$ and $\alpha, \beta
    \ge 0$
    \[ \prob(\Delta_t^b \ge \Delta_t^a) \le \exp\left(-qt\left(1-2q_b^{\alpha/(\alpha + \beta)} q_a^{\beta/(\alpha + \beta)}\right)\right) \]
\end{lemma}
\begin{proof}
Let $q := p_a(1-p_b)+(1-p_a)p_b$ be the probability that exactly one process
makes progress in a single time step. Let $q_a := p_a(1-p_b)/q$ be the
conditional probability of $\mathcal{A}$ making progress, given that one
process makes progress, and define $q_b$ likewise. Assume that in $t$ steps we
have $\ell$ progressing steps. Then the probability that $\mathcal{B}$
    makes at least as much progress as $\mathcal{A}$ is $\prob(\Bin(\ell, q_b) \ge \lceil \ell\alpha/(\alpha + \beta) \rceil)$. Then,
\begin{equation}
    \label{eq:prob-progress-b-better-than-a}\prob(\Delta_t^b \ge \Delta_t^a) = \sum_{\ell=0}^t \prob(\Bin(t, q) = \ell) \cdot \prob(\Bin(\ell, q_b) \ge \lceil \ell\alpha/(\alpha + \beta) \rceil)
\end{equation}

Note that $p_b \le p_a$ is equivalent to $q_b \le q_a$. Thus, $q_b/q_a \le 1$. Hence

\begin{align*}
    &\prob(\Bin(\ell, q_b) \ge \lceil \ell\alpha/(\alpha + \beta) \rceil) =
    \sum_{i=\lceil \ell\alpha/(\alpha + \beta) \rceil}^\ell \binom{\ell}{i} q_b^{i} q_a^{\ell-i}\\
        =\;& \sum_{i=\lceil \ell\alpha/(\alpha + \beta) \rceil}^\ell \binom{\ell}{i} q_b^{\ell\alpha/(\alpha + \beta)} q_a^{\ell - (\ell\alpha/(\alpha + \beta))} (q_b/q_a)^{i - (\ell\alpha/(\alpha + \beta))} \\
        \le\;& 2^\ell q_b^{\ell\alpha/(\alpha + \beta)} q_a^{\ell - (\ell\alpha/(\alpha + \beta))}
        = \left(2 q_b^{\alpha/(\alpha + \beta)} q_a^{\beta/(\alpha + \beta)} \right)^\ell.
\end{align*}

    Using the above in~\eqref{eq:prob-progress-b-better-than-a} and $\prob(\Bin(t, q) = \ell) = \binom{t}{\ell} q^\ell (1-q)^{t-\ell}$ yields,

\begin{align*}
    \prob(\Delta_t^b \ge \Delta_t^a) \le\;&
    \sum_{\ell=0}^t \binom{t}{\ell} q^\ell (1-q)^{t-\ell}  \cdot \left(2 q_b^{\alpha/(\alpha + \beta)} q_a^{\beta/(\alpha + \beta)} \right)^\ell \\
    =\;& \sum_{\ell=0}^t \binom{t}{\ell} (1-q)^{t-\ell}  \cdot \left(2q \cdot q_b^{\alpha/(\alpha + \beta)} q_a^{\beta/(\alpha + \beta)} \right)^\ell \\
    \intertext{(using the Binomial Theorem)}
    =\;& \left(1-q+ 2q \cdot q_b^{\alpha/(\alpha + \beta)} q_a^{\beta/(\alpha + \beta)} \right)^t\\
    =\;& \left(1-q\left(1-2q_b^{\alpha/(\alpha + \beta)} q_a^{\beta/(\alpha + \beta)}\right)\right)^t\\
    \le\;& \exp\left(-qt\left(1-2q_b^{\alpha/(\alpha + \beta)} q_a^{\beta/(\alpha + \beta)}\right)\right).\qedhere
    \end{align*}
\end{proof}

Applying this lemma allows us to derive a lower bound on the probability that
$RLS_a$ wins against RLS$_b$ ($a < b$) with a cutoff time of $\kappa$. Additional arguments for small $\kappa/\binom{n}{a}$ show that 
the probability that RLS$_a$ wins is always at least $1/2$.
\begin{lemma}
    \label{lem:min_t_a_ahead_b}
    For every $1 \le a < b = o(n)$, in an evaluation with a single run on
    {\scshape Ridge*} with cutoff time $\kappa$, RLS$_a$ wins against RLS$_b$
    with probability at least
    \[
        \max\left\{\frac{1}{2}, \  1 - \exp  \left( -\kappa/\binom{n}{a} \cdot (1-o(1))\right) - \exp(-\Omega(n/b)) \right\}
    \]
\end{lemma}

%% file: sections/ridge_tuner_analysis.tex
\subsection{{\scshape ParamRLS-F} Performance Analysis}
\label{sec:tuner_ridge_analysis}

Using the above lemmas, we now consider the cutoff time required before
ParamRLS returns $k=1$ in expectation.
The following theorem shows that one run per configuration evaluation suffices for large enough cutoff times.
Note that it is not sufficient for the
active parameter merely to be set to the value 1, since it is still possible
for it to then change again to a different value. We therefore require that the
active parameter remains at 1 for the remainder of the tuning time. We
calculate this probability in the same theorem.  \par

\begin{theorem}
    \label{thm:ridge_tuning_time}
    ParamRLS-F for RLS$_k$ on {\scshape Ridge*} with $\phi \le \sqrt{n}$, cutoff time $\kappa = \omega(n)$, local search operator $\pm 1$ and
    any initial parameter value, in expectation after at most
    $2\phi^{2}$ evaluations with a single run each has
    active parameter $k=1$. If ParamRLS-F runs for $T \ge 4\phi^{2}$ evaluations, then it
    returns the parameter value $k=1$ with probability at least $1 - 2^{-\Omega(T/\phi^2)} - T \cdot (2^{-\Omega(\kappa/n)} + 2^{-\Omega(n)})$.
\end{theorem}

\begin{proof}
    By Lemma~\ref{lem:min_t_a_ahead_b}, the probability that RLS$_a$ beats
    RLS$_b$ in an evaluation with any cutoff time is at least $1/2$.  We can
    therefore model the tuning process as the value of the active parameter
    performing a lazy random walk over the states $1, \ldots, \phi$. We
    pessimistically assume that the active parameter
    decreases and increases by $1$ with respective probabilities $1/4$ and that it stays the same with probability $1/2$.

    Using standard random walk arguments~\cite{Feller1968,Feller1971}, the expected first hitting time of state~1 is at most $2\phi^{2}$. By Markov's
    inequality, the probability that state~1 has not been reached in $4\phi^{2}$ steps is at most $1/2$. Hence the probability that state~1 is not reached during $\lfloor T/4\phi^{2} \rfloor$ periods each consisting of $4\phi^{2}$ steps is $2^{-\lfloor T/4\phi^{2} \rfloor} = 2^{-\Omega(T/\phi^{2})}$.

    Once state~1 is reached, we remain there unless RLS$_2$ beats RLS$_1$ in a run. By Lemma~\ref{lem:min_t_a_ahead_b}, this event happens in a specific evaluation with probability at most $2^{-\Omega(\kappa/n)} + 2^{-\Omega(n)}$. By a union bound over at most $T$ evaluations, the probability that this ever happens is at most $T \cdot (2^{-\Omega(\kappa/n)} + 2^{-\Omega(n)})$.
\end{proof}

We now show that even for extremely small cutoff times i.e., $\kappa=1$, the algorithm can identify the correct configuration as long as sufficient number of runs are executed per configuration evaluation.
\begin{theorem}
    \label{thm:can_tune_for_ridge_cutoff_time_1}
    Consider ParamRLS-F for RLS$_k$ on {\scshape Ridge*} with $T$ evaluations, each consisting of $n^{3/2}$
    runs with cutoff time $\kappa = 1$. Assume we are using the local search operator $\pm 1$. In expectation the tuner requires
    at most $2\phi^{2}$ evaluations in order to set the active parameter to
    $k=1$. If the tuner is run for $T \ge 4\phi^{2}$ evaluations then
    it returns the value $k=1$ with probability at least
    \[
        1 - 2^{-\Omega(T/\phi^2)} - T \cdot (2^{-\Omega(\kappa/n)} + 2^{-\Omega(n)}).
    \]
\end{theorem}

\begin{proof}
    Define $X_{r}$ as the number of runs out of $r$ runs, each with cutoff time
    $\kappa = 1$, in which RLS$_1$ makes progress.  Define $Y_{r}$ as the
    corresponding variable for RLS$_2$.  Let $T = n^{3/2}$. By Chernoff bounds,
    we can show that $P(X_{r} > \sqrt{n}/2) \ge 1 - \exp(-\Omega(\sqrt{n}))$.
    We can also show that, again by Chernoff bounds, $P(Y_{r} < \sqrt{n}/2) \ge
    1 - \exp(-\Omega(\sqrt{n}))$. Therefore, with overwhelming probability, RLS$_1$
    has made progress in more of these $n^{3/2}$ runs than RLS$_2$. That is,
    with overwhelming probability, RLS$_1$ wins the evaluation. \par

    It is easy to show that, for $a<b$, RLS$_a$ beats RLS$_b$ with probability
    at least $1/2$.  This means that we can make the same pessimistic
    assumption about the progress of the value of the active parameter as we do
    in the proof of Theorem~\ref{thm:ridge_tuning_time}.  The remainder of the
    proof is identical.
\end{proof}

\subsection{{\scshape ParamRLS-T} Performance Analysis}

We conclude the section by showing that,  unless the cutoff time is large, ParamRLS-T returns a value of $k$ chosen uniformly at random for RLS$_k$ and {\scshape Ridge*}.
\begin{theorem}
    Consider ParamRLS-T for RLS$_k$ on {\scshape Ridge*} with $\phi \le \sqrt{n}$, local search operator $\pm 1$ or $\pm\{1,2\}$, cutoff time $\kappa \le
    n^{2}/2$, and $T$ evaluations consisting each of $r$ runs. With
    overwhelming probability, for any polynomial choices of $T$ and $r$, the
    tuner will return a value for $k$ chosen uniformly at random.
    \label{thm:pRLST_no_good_ridge_small_cutoff}
\end{theorem}

\begin{proof}
    For all $k \le \sqrt{n}$, we have $\kappa \le {n \choose k} \lfloor n/k \rfloor / 2$.
    By Lemma~\ref{lem:opt_lower_bound} with probability at least $1 - \exp(-\Omega(n))$, no RLS$_k$ with $k \le \sqrt{n}$ will have reached the optimum of
    {\scshape Ridge*} within $\kappa$ iterations. Thus,
    with probability at least $1 - r \cdot T \cdot (\exp(-\Omega(n)))$, no configuration reached the optimum of {\scshape Ridge*}
    in any of the $r$ runs in any of the $T$ evaluations. In this case, we
    can simply use the random walk argument as used in the proof of
    Theorem~\ref{thm:ridge_tuning_time}, but in this case the value of the
    active parameter will not settle on $k=1$, meaning that ParamRLS-T will
    return a value for $k$ chosen uniformly at random.
\end{proof}

%% file: sections/rlsk_om_analysis.tex
In this section we analyse the performance of ParamRLS when configuring RLS$_k$ for {\scshape OneMax}.
If RLS$_k$ is only allowed to run for few fitness function evaluations, then the algorithm with larger parameter values for $k$ performs better than with smaller ones.
On the other hand, if  more fitness evaluations are allowed, then  RLS$_1$ will be the fastest at identifying the optimum \cite{DoerrYangArxiv}.
Our aim is to show that ParamRLS-F can identify whether $k=1$ is the optimal parameter choice or whether a larger value for $k$ performs better according to whether the cutoff time is small or large.
Hence, to prove our point it suffices to consider the configurator with the following parameter vector: $k \in [1,2,3,4,5]$ which also simplifies the analysis.
We will prove that ParamRLS-F identifies that $k=1$ is optimal for any $\kappa \ge 4n$ even for single runs per configuration evaluation.
This time is shorter than the expected time required by any configuration to optimise  {\scshape OneMax} (i.e., $\Theta(n \ln n)$) \cite{Lehre2012}.
If, instead, the cutoff time is smaller than $0.03n$, then ParamRLS-F will identify that $k>1$ is a better choice, as desired.


%

The following lemma gives bounds on the expected progress towards the optimum in one step.
\begin{lemma}
\label{lem:drift-bounds}
The expected progress $\Delta_k(s)$ of RLS$_k$ with current distance~$s$ to the optimum is
\[
\Delta_k(s) = \sum_{i=\lfloor k/2 \rfloor +1}^k (2i-k) \cdot \binom{s}{i}\binom{n-s}{k-i}/\binom{n}{k}
\]
In particular, for $s \ge k$,
\begin{align*}
\Delta_1(s) =\;& \frac{s}{n}\\
\Delta_2(s) =\;& \frac{2s(s-1)}{n(n-1)} \le 2\left(\frac{s}{n}\right)^2 \quad \Delta_3(s) = \frac{3s(s-1)}{n(n-1)} \le 3\left(\frac{s}{n}\right)^2\\
\Delta_4(s) =\;& \frac{8s(s-1)(s-2)(n-s/2-3/2)}{n(n-1)(n-2)(n-3)} \le 8\left(\frac{s}{n}\right)^3\\
\Delta_5(s) =\;& \frac{10s(s-1)(s-2)(n-s/2-3/2)}{n(n-1)(n-2)(n-3)} \le 10\left(\frac{s}{n}\right)^3.
\end{align*}
\end{lemma}

It is well known that RLS$_1$ has the lowest expected optimisation time on
{\scshape OneMax} for all RLS$_k$. It runs in expected time $n \ln n \pm O(n)$, which is best possible for all unary unbiased black-box algorithms~\cite{Doerr:2016:OPC:2908812.2908950,DoerrYangArxiv} up to terms of $\pm O(n)$.
It is also known~\cite{Doerr:2016:OPC:2908812.2908950,DoerrYangArxiv} that,
regardless of the fitness of the individual, flipping $2c$ bits never gives
higher expected drift than flipping $2c+1$ bits (for any positive integer
$c$). For this reason, we use the local search operator $\pm \{1, 2\}$.

%% file: sections/om_long_races.tex
\subsection{\boldmath$k=1$ is Optimal for Large Cutoff Times}

For large cutoff times, ParamRLS-F is able to identify the optimal parameter value $k=1$. The analysis is surprisingly challenging as most existing methods in the runtime analysis of evolutionary algorithms are geared towards first hitting times. Results on the expected fitness after a given cutoff time (fixed-budget results) are rare~\cite{paper:fixed_budget_analysis,paper:fixed_budget_linear_funcs,Doerr2013c,Jansen2014,Nallaperuma2017} and do not cover RLS$_k$ for $k > 1$.

The following lemma establishes intervals $[\ell_i, u_i]$ such that the current distance to the optimum is contained in these intervals with overwhelming probability.
\begin{lemma}
\label{lem:confidence-intervals}
Consider RLS$_k$ on \textsc{OneMax} with a cutoff time $\kappa \ge 4n$. Divide the first $4n$ generations into 80 periods of length $n/20$ each.
Define
$\ell_0 = n/2 - n^{3/4}$ and $u_0 = n/2 + n^{3/4}$ and, for all $1 \le i \le 80$,
\begin{align*}
    \ell_i =\;& \ell_{i-1} - \frac{n}{20} \Delta_k(\ell_{i-1}) - o(n) \text{\enskip and \enskip} u_i = u_{i-1} - \frac{n}{20}\Delta_k(\ell_i) + o(n).
\end{align*}
    Then, with overwhelming probability at the end of period~$i$ for $0 \le i \le 80$, the current distance to the optimum is in the interval $[\ell_i, u_i]$ and throughout period~$i$, $1 \le i \le 80$, it is in the interval $[u_{i-1}, \ell_i]$.
\end{lemma}
\begin{proof}
We prove the statement by induction. At time~0, the current distance to the optimum is in $[n/2-n^{3/4}, n/2+n^{3/4}]$ with overwhelming probability by Chernoff bounds.

Now assume that at the end of period~$i-1$, the current distance is in $[\ell_{i-1}, u_{i-1}]$.
In order to determine the next lower bound $\ell_i$ on the distance, we temporarily assume that at the end of period~$i-1$, we are precisely at distance~$\ell_{i-1}$. This assumption is pessimistic here since starting period~$i$ closer to the optimum can only decrease the distance to the optimum at the end of period~$i$.

During period~$i$, since the current distance can only decrease and the expected progress is non-decreasing in the distance, the expected progress in each step is at most $\Delta_k(\ell_{i-1})$. By the method of bounded martingale differences~\cite[Theorem 3.67]{paper:scheideler_hab_thesis}, the total progress in $n/20$ steps is thus at most $n/20 \cdot \Delta_k(\ell_{i-1}) + (n/20)^{3/4} = n/20 \cdot \Delta_k(\ell_{i-1}) + o(n)$ with probability
\[
1-\exp\left(-((n/20)^{3/4})^2/(2k^2 n/20)\right) = 1 - \exp(\Omega(-n^{1/2})).
\]
Hence we obtain $\ell_i = \ell_{i-1} - \frac{n}{20} \Delta_k(\ell_{i-1}) - o(n)$ as a lower bound on the distance at the end of period~$i$, with overwhelming probability.

While the distance in period~$i$ is at least $\ell_i$, the expected progress in every step is at least $\Delta_k(\ell_i)$.
    Again using the method of bounded differences, by the same calculations as above, the progress is at least $n/20 \cdot \Delta_k(\ell_i) - o(n)$ with overwhelming probability. This establishes $u_i = u_{i-1} - n/20 \cdot \Delta_k(\ell_i) + o(n)$ as an upper bound on the distance at the end of period~$i$.
Taking the union bound over all failure probabilities proves the claim.
\end{proof}

Iterating the recurrent formulas from Lemma~\ref{lem:confidence-intervals} shows the following.
\begin{lemma}
\label{lem:numerical-distances}
After $4n$ steps, \wop RLS$_1$ is ahead of RLS$_2$ and RLS$_3$ by a linear distance:
$u_{80,1} \le \ell_{80, 2} - \Omega(n)$
and
    $u_{80,1} \le \ell_{80, 3} - \Omega(n)$ respectively.
Furthermore, \wop RLS$_3$ is ahead of RLS$_4$ and RLS$_5$ by a linear distance:
$u_{80,3} \le \ell_{80, 4} - \Omega(n)$
and
$u_{80,3} \le \ell_{80, 5} - \Omega(n)$ respectively.
And \wop the distance to the optimum is at most $0.17n$ for RLS$_1$, RLS$_3$ and RLS$_5$.
\end{lemma}

We conclude that for every $\kappa \ge 4n$, smaller parameters win with overwhelming probability.
\begin{theorem}
For every cutoff time $\kappa \ge 4n$, with overwhelming probability RLS$_1$ beats RLS$_2$ as well as RLS$_3$ and RLS$_3$ beats RLS$_4$ as well as RLS$_5$.
    \label{thm:small_k_wins_long_runs_om}
\end{theorem}
\begin{proof}
%
Lemma~\ref{lem:numerical-distances}
proves the claim for a cutoff time of $\kappa = 4n$. For larger cutoff times, it is possible for the algorithms that lag behind to catch up.
To this end, we define the distance between two algorithms RLS$_a$, RLS$_b$ with $a < b$ as $D_t^{a,b} := s_{t, b} - s_{t, a}$, where $s_{t, a}$ and $s_{t,b}$ refer to the respective distances to the optimum at time~$t$. Initially we have $D_t^{a,b} = \Omega(n)$ for all considered algorithm pairs. We then show that, as long as $D_{t}^{a,b} \le n^{1/4}$, the distance has a tendency to increase. We then apply the negative drift theorem~\cite{Oliveto2011,Oliveto2012Erratum} in the version for self-loops~\cite{Rowe2013} to show that with overwhelming probability $D_t^{a,b}$ does not drop to 0 until RLS$_a$ has found an optimum ($s_{t,a} < a$).
Details are omitted due to space restrictions.
%
\end{proof}

We are now able to derive the expected number of evaluations required for the
tuner to return $k=1$ for RLS$_k$ on {\scshape OneMax} with a large enough
cutoff time (for these results to hold, we assume that we use a local search
operator of $\pm \{1,2\}$).

\begin{theorem}
    For ParamRLS-F tuning RLS$_k$ for {\scshape OneMax}, with cutoff time
    $\kappa \ge 4n$, $\phi = 5$, local search operator $\pm \{1,2\}$, $T$ evaluations and $r$ runs per evaluation, with $T$ and
    $r$ both polynomial, then in expectation we require at most 8 evaluations
    before the active parameter is set to $k=1$ for the first time. 
    If $T = \Omega(n^{\varepsilon})$ for some constant $\varepsilon > 0$ then the tuner returns the parameter $k=1$ \wop
\end{theorem}

\begin{proof}
    We use a similar technique to that used in the proof of
    Theorem~\ref{thm:ridge_tuning_time}. In this case, however, we split the
    state space of the value of the active parameter into just three states:
    $(1)$, $(2,3)$, and $(4,5)$.
    We know from
    Theorem~\ref{thm:small_k_wins_long_runs_om} that RLS$_3$ beats RLS$_4$ and
    RLS$_5$ with overwhelming probability in a run with cutoff time $\kappa \ge
    4n$. Let us assume that this always happens. Then the transition
    probability from state $(4,5)$ to state $(2,3)$ is at least $1/4$, since this is the
    probability that we evaluate RLS$_5$ against RLS$_3$ or RLS$_4$ against
    RLS$_3$. In all other cases, depending on whether RLS$_4$ beats RLS$_2$, we either move to state $(2, 3)$ or stay in state $(4,5)$. By a similar argument, the
    transition probability from state $(2,3)$ to state $(1)$ is at least $1/4$, and with
    probability at most $3/4$ we remain in state $(2,3)$. Therefore, in the worst case
    (where the initial choice for the parameter $k$ puts us in state $(4,5)$),
    we require, in expectation, at most 8 evaluations before we hit state $(1)$. \par

    A Chernoff bound for geometric random variables \cite[Theorem~1.14]{chapter:doerr_tools_from_prob_theory} tells us that the
    probability that we require more than $T$ evaluations to hit state $(1)$
    when starting from state $(4,5)$ is at most $\exp(-(T-8)^{2}/(16T)) =
    \exp(-\Omega(T))$. If $T = \Omega(n^{\varepsilon})$ for some constant $\varepsilon > 0$ then \wop $T$ evaluations are sufficient.
    Recall that we still need the probability that we remain
    in state $(1)$ after hitting it for the first time. In the worst case, this
    means that we require that RLS$_1$ beats RLS$_2$ or RLS$_3$ for all $T
    \cdot r$ runs within the tuning process. Recall that \wop RLS$_1$ beats RLS$_2$ and \wop RLS$_1$ beats RLS$_3$. By
    Theorem~\ref{thm:small_k_wins_long_runs_om} and the definition of overwhelming probabilities, the
    probability that we remain in state $(1)$ after hitting it for the first
    time is therefore at least $1 - T \cdot r \cdot \exp(-\Omega(n^{\varepsilon'}))$ for some constant $\varepsilon' > 0$.
\end{proof}

%% file: sections/om_short_races.tex
\subsection{\boldmath$k>1$ is Optimal for Small Cutoff Times}
We now show that if the cutoff time is small, then ParamRLS-F identifies that $k=1$ is not optimal anymore as desired.



\begin{lemma}
    \label{lem:rls1_loses_vs_rls2_rls3_small_kappa}
    For cutoff time $\kappa \le 0.03n$ the probability that RLS$_1$ beats RLS$_3$ is at most $4e^{-\Omega(\kappa^2/n)} + e^{-\Omega(\kappa)}$.
    \label{lem:RLS3_beats_RLS1_short_race_wop}
    The same holds for the probability that RLS$_2$ beats RLS$_3$. \footnote{Note that the result is only meaningful for $\kappa = \Omega(\sqrt{n})$ as otherwise we get a trivial probability bound of $4e^{-\Omega(\kappa^2/n)} \ge 1$}
\end{lemma}

\begin{proof}
Let $s_{t,1}$ be the distance to the optimum in RLS$_1$ and $s_{t,3}$ be the distance to the optimum in RLS$_3$ at time~$t$.
Let $\varepsilon > 0$ be a constant chosen later, then by Chernoff bounds,
    \[ \prob(s_{0,1}, s_{0,3} \in [(n-\varepsilon\kappa) / 2, (n+\varepsilon\kappa) / 2]) \ge 1 - 4e^{-\Omega(\kappa^2/n)} \]
We assume in the following that this is the case.
Then RLS$_3$ wins if in $\kappa$ steps RLS$_3$'s progress exceeds that of
RLS$_1$ by at least $\varepsilon \kappa$.

%
Define $D_t := (s_{t+1,3}-s_{t,3})-(s_{t+1,1}-s_{t,1})$ to be the difference in the progress values made by the two algorithms.
Along with the drift bounds from Lemma~\ref{lem:drift-bounds},
\[
\E(D_t) = \frac{s_{t,3}}{n} \cdot \frac{3(s_{t,3}-1)}{n-1} - \frac{s_{t,1}}{n}
= 3(s_{t,3}/n)^2 - s_{t,1}/n - O(1/n).
\]
Note that the leading constant in $\kappa$ is chosen as $0.03 < \gamma := 1/3 \cdot (1/2-1/\sqrt{6})$. This implies that for $t \le 0.03n$ we always have $s_{t,1} \le n/2 + \varepsilon \kappa \le n/2 + \varepsilon n$ and $s_{t,3} \ge n/2-\varepsilon \kappa -0.09n$.
    We bound the latter using $\varepsilon \kappa \le \varepsilon n$ and $0.09n = 3\gamma n - 3(\gamma-0.03)n \le 3\gamma n - 2\varepsilon n$ if we choose $\varepsilon$ small enough, we have $s_{t,3} \ge n/2 - \varepsilon n - (1/2-1/\sqrt{6})n + 2\varepsilon n =  n/\sqrt{6} + \varepsilon n$.
Using these inequalities,
\begin{align*}
\E(D_t) \ge\;& 3(1/\sqrt{6} + \varepsilon)^2 - (1/2+\varepsilon) - O(1/n)\\
=\;& 1/2 + \sqrt{6}\varepsilon + 3\varepsilon^2 - 1/2 - \varepsilon  - O(1/n)
\ge (\sqrt{6}-1)\varepsilon - O(1/n).
\end{align*}
    Now, for $D := \sum_{t=1}^\kappa D_t$, using $\E(D)  \ge \varepsilon \kappa + (\sqrt{6}-2)\varepsilon \kappa - O(\kappa/n) = \varepsilon \kappa + \Omega(\kappa)$ we derive $\prob(D \le \varepsilon \kappa) \le \prob(D \le \E(D) - \Omega(\kappa))$. By the method of bounded differences~\cite[Theorem 3.67]{paper:scheideler_hab_thesis}, this is at most
$    \exp(-\Omega(\kappa^2)/\Theta(\kappa))
    = \exp(-\Omega(\kappa))$.
\end{proof}


\begin{theorem}
    \label{thm:k_ge1_returned_for_small_cutoff}
    When tuning RLS$_k$ for {\scshape OneMax}, the probability that ParamRLS-F
    with cutoff time $\kappa \le 0.03n$, local search operator $\pm 1$ or $\pm
    \{1,2\}$ and $\phi=5$ returns the value $k=1$, for any number of
    evaluations $T$, is at most $T \cdot(4e^{-\Omega(\kappa^2/n)} +
    e^{-\Omega(\kappa)})$.
\end{theorem}

\begin{proof}
    In order for ParamRLS-F to return a value of $k=1$, it is necessary for
    RLS$_1$ to beat either RLS$_2$ or RLS$_3$ in at least one evaluation. In the best
    case scenario, each evaluation in the tuning process will be either RLS$_2$ or
    RLS$_3$ against RLS$_1$, since this maximises the number of opportunities in
    which RLS$_1$ has to win one of these evaluations. Using the upper bounds on the
    probabilities of RLS$_1$ beating RLS$_2$ and RLS$_3$ (see
    Lemma~\ref{lem:RLS3_beats_RLS1_short_race_wop}), the union bound tells us that the probability that RLS$_1$
    wins any one of these $T$ evaluations is at most
    $T \cdot(4e^{-\Omega(\kappa^2/n)} + e^{-\Omega(\kappa)})$.
\end{proof}

%% file: sections/conclusion.tex
\section{Conclusions}
We have shown that the cutoff time only slightly impacts the performance of ParamRLS-F.
ParamRLS-F can identify that $k=1$ is the optimal parameter value for both optimisation problems for large enough cutoff times.
Surprisingly, for such cutoff times, a single run per configuration evaluation is sufficient to achieve the desired results.
While we do not expect this to be the case for harder optimisation problems, it is promising that for the simple unimodal problems considered herein
multiple configuration evaluations are not necessary.
Furthermore the required cutoff times of $\kappa=\omega(n)$ and $\kappa=4n$, respectively for {\scshape Ridge*} and {\scshape OneMax},
are considerably smaller than the expected time for any parameter configuration to optimise either problem (i.e., $\Omega(n^2)$ and $\Omega(n \log n)$ respectively for the best configuration ($k=1$)).
On the other hand, if the cutoff times are small ParamRLS-F identifies that for {\scshape Ridge*} the optimal parameter value is still $k=1$ as long as sufficient runs are performed to evaluate the performance of parameter configurations.
We prove this effect for the extreme value $\kappa=1$ for which $n^{3/2}$ runs suffice to always identify the better configuration w.o.p.
Note that $n^{3/2}$ runs lasting one generation each  are still considerably smaller than the time required for any configuration to identify the optimum of {\scshape Ridge*}.
Concerning {\scshape OneMax}, instead, for cutoff times smaller than $\kappa=0.03n$ we proved that ParamRLS-F identifies that $k=1$ is not the best parameter, 
as desired
(i.e., RLS$_3$ will produce better solutions than RLS$_1$ if the time budget is small).

The impact of the cutoff time on ParamRLS-T, instead, is very big.
The configurator cannot optimise the single parameter of RLS$_k$ applied to any function, even functions with up to exponentially many optima, if the cutoff time is smaller than $\kappa=(n \ln n)/2$
independent of the number of runs per configuration evaluation. For small cutoff times, even if the tuner happens to set the active parameter to the optimal value, it will not be identified as optimal, making it unlikely that it stays there for the remainder of the tuning process.
For the unimodal  {\scshape Ridge*} function at least a quadratic cutoff time is required.

%% file: sections/appendix.tex
\clearpage

\appendix

\section{Proofs Omitted from the Main Part}

This appendix contains proofs omitted from the main part of the paper due to space restrictions.

\subsection{Proof of Lemma~\ref{lem:min_t_a_ahead_b}}

\begin{proof}
Using the notation from Lemma~\ref{lem:const_probs_diff_bound}, we have
$p_{a} = 1/{n \choose a}$ and $p_{b} = 1/{n \choose b}$, which implies $p_b = o(p_a)$ since $b = o(n)$.
Further, $q \ge 1/\binom{n}{a} \cdot (1-o(1))$, $q_a = 1-o(1)$ and $q_b = p_b(1-p_a)/q \le p_b(1-p_a)/(p_a(1-p_b) \le p_b/p_a = \frac{b!(n-b)!}{a!(n-a)!} \le (b/(n-b))^{b-a}$. This implies $q_b^{a/(a+b)} \le (b/(n-b))^{a(b-a)/(a+b)}$.
Using $b/(n-b) = o(n)/n = o(1)$ and $a(b-a)/(a+b) \ge a/(2a+1) \ge 1/3$, we obtain $q_b^{a/(a+b)} = o(1)$.
    Lemma~\ref{lem:const_probs_diff_bound} tells us that RLS$_a$ is ahead of
    RLS$_b$ with probability at least
    \[
        1 - \exp  \left( -\kappa/\binom{n}{a} \cdot (1-o(1))\right).
    \]
    The above argument ignores that progress stops once a global optimum is reached. If RLS$_a$ reaches a global optimum and RLS$_b$ does not, RLS$_a$ still wins. We use the union bound to include a term reflecting the possibility that RLS$_b$ finds the global optimum. By Lemma~\ref{lem:opt_lower_bound}, if $\kappa \le \binom{n}{b}\lfloor n/b \rfloor/2$, the probability that RLS$_b$ does find the optimum is at most $\exp(-\lfloor n/b \rfloor /6)$.
    This proves the claimed bound for $\kappa \le \binom{n}{b}\lfloor n/b \rfloor/2$.

    For larger $\kappa$ we argue that by Lemma~\ref{lem:opt_lower_bound}, the probability that RLS$_a$ finishes within the first $\binom{n}{b}\lfloor n/b \rfloor/2 \ge 2(\binom{n}{a} \lfloor n/a \rfloor)$ steps is $1-\exp(-\Omega(n/a)) \ge 1- \exp(-\Omega(n/b))$. Along with the fact that RLS$_b$ with the same probability needs more than $\binom{n}{b}\lfloor n/b \rfloor/2$ steps, this proves that RLS$_a$ wins with probability at least $1-\exp(-\Omega(n/b))$ for $\kappa > \binom{n}{b}\lfloor n/b \rfloor/2$.

We have proved the claim for all $\kappa \ge 2\binom{n}{a}$, assuming $n$ is large enough to make the $o(1)$-term less than $1/2$.
For $\kappa < 2\binom{n}{a}$ we argue that RLS$_b$ can only win if it makes progress in $\kappa$ steps. The probability for this is at most $\kappa/\binom{n}{b}$. RLS$_a$ wins for sure if it does make progress in $\kappa$ steps and RLS$_b$ does not make progress. The probabilities for these events are at least $1-\left(1-1/\binom{n}{a}\right)^\kappa \ge \kappa/(\kappa+\binom{n}{a})$ (using $1-(1-p)^\lambda \ge p\lambda/(1+p\lambda)$~\cite[Lemma~6]{Badkobeh2015}) and $1-\kappa/\binom{n}{b} = 1 - o(1)$, respectively. So the probability that they both occur is at least
\[
\frac{\kappa}{\kappa+\binom{n}{a}} \cdot (1-o(1)) \ge \frac{\kappa}{3\binom{n}{a}} \cdot (1-o(1)) > \frac{\kappa}{\binom{n}{b}}
\]
for large enough~$n$. Hence, in all cases where at least one algorithm makes progress, RLS$_a$ is more likely to win than RLS$_b$. In all other cases there is a tie and the probability that RLS$_a$ is declared winner is $1/2$. This proves a lower bound of $1/2$ for the probability that RLS$_a$ wins.
\end{proof}

\subsection{Full proof of Lemma~\ref{thm:can_tune_for_ridge_cutoff_time_1}}

\begin{proof}
    Define $X_{r}$ as the number of runs out of $r$ runs, each with cutoff time
    $\kappa = 1$, in which RLS$_1$ makes progress.  Define $Y_{r}$ as the
    corresponding variable for RLS$_2$.  Let $T = n^{3/2}$. By Chernoff bounds,
    we can show that $P(X_{r} > \sqrt{n}/2) \ge 1 - \exp(-\Omega(\sqrt{n}))$.
    We can also show that, again by Chernoff bounds, $P(Y_{r} < \sqrt{n}/2) \ge
    1 - \exp(-\Omega(n))$. Therefore, with overwhelming probability, RLS$_1$
    has made progress in more of these $n^{3/2}$ runs than RLS$_2$. That is,
    with overwhelming probability, RLS$_1$ wins the evaluation. \par

    We can analyse this tuning process as a whole in the same way in which we
    analyse the tuning process in the proof of
    Theorem~\ref{thm:ridge_tuning_time}. We first observe that, in order for
    RLS$_a$ to beat RLS$_b$ (with $a<b$) in a run with cutoff time $\kappa =
    1$, it is sufficient for it to have made a leap and for RLS$_b$ to have
    failed to do so. Letting $A$ be the event that RLS$_a$ beats RLS$_b$ in a
    run with cutoff time $\kappa = 1$, we have
    \[ \prob(A) \ge \frac{1}{{n \choose a}} \left( 1 - \frac{1}{{n \choose b}} \right) \]
    Let $B$ denote the event that RLS$_b$ beats RLS$_a$ in a run with cutoff time $\kappa=1$. Since RLS$_b$ making
    progress is a necessary condition for event $B$ to take place, we have
    $\prob(B) \le 1/{n \choose b}$. For large enough $n$, we have that
    \[ \frac{1}{{n \choose a}} \left( 1 - \frac{1}{{n \choose b}} \right) \ge 1/{n \choose b} \]
    which implies that $P(A) \ge P(B)$. This means that, for any $1 \le x \le r$ the probability that
    RLS$_a$ wins $x$ runs in an evaluation is at least the probability that RLS$_b$ wins $x$
    runs. Observing that if an evaluation does not end in a draw then the
    winner must have won more runs than its competitor, we see that, since,
    $P(A) \ge P(B)$, the winner must be RLS$_a$ with probability at least $1/2$.
    This means that we can make the same pessimistic assumption as we do in the
    proof of Theorem~\ref{thm:ridge_tuning_time}. The remainder of the proof is
    identical.
\end{proof}

\subsection{Proof of Lemma~\ref{lem:drift-bounds}}

\begin{proof}
We first compute the probability of
flipping a certain number of bits in a bit string using RLS$_k$.
If the bit string currently has Hamming distance $s$ to the optimum, then the probability that a $k$-bit mutation flips
    exactly $i$ bits that disagree with the optimum and $k-i$ bits that agree with the optimum is
    \begin{equation}
    \label{eq:hypergeometric}
     {s \choose i} {n-s \choose k-i}/{n \choose k}
     \end{equation}
This corresponds to a hypergeometric distribution with parameters $s$ and $n$.

If a $k$-bit mutation flips $i$ disagreeing bits and $k-i$ agreeing bits, the distance to the optimum decreases by $i-(k-i)=2i-k$. This is only accepted if $2i-k \ge 0$, and progress is only made if $2i-k > 0$ or, equivalently, $i > \lfloor k/2\rfloor$. The claim then follows from~\eqref{eq:hypergeometric} and the definition of the expectation.

By~\cite[Lemma~27]{DoerrYangArxiv} we have $\Delta_2(s) = 2\Delta_3(s)/3$ and $\Delta_4(s)= 4\Delta_5(s)/5$, hence we only need to show the claims for $\Delta_1(s), \Delta_3(s)$, and $\Delta_5(s)$.
The formula $\Delta_1(s) = s/n$ follows immediately. For $\Delta_3(s)$ we have
\begin{align*}
\Delta_3(s) =\;& \left(\binom{s}{2}\binom{n-s}{1} + 3\binom{s}{3}\binom{n-s}{0}\right)/\binom{n}{3}\\
=\;& \left(\frac{s(s-1)(n-s)}{2} + \frac{3s(s-1)(s-2)}{6}\right)/\binom{n}{3}\\
=\;& \left(\frac{s(s-1)(n-2)}{2}\right)/\binom{n}{3}
= \frac{3s(s-1)}{n(n-1)}
\end{align*}

For $\Delta_5(s)$ we have
    \begin{align*}
    \Delta_5(s) &=\; \left(\binom{s}{3}\binom{n-s}{2} + 3\binom{s}{4}\binom{n-s}{1} + 5\binom{s}{5}\binom{n-s}{0} \right)/\binom{n}{5} \\
        &= \left(\left(\frac{s(s-1)(s-2)}{6}\right) \left( \frac{(n-s)(n-s-1)}{2} \right) \right. \\
        &\quad\quad+ \frac{3s(s-1)(s-2)(s-3)(n-s)}{24} \\
        &\quad\quad+ \left. \frac{5s(s-1)(s-2)(s-3)(s-4)}{120} \right) \\
        &\quad\quad/ \left( \frac{n(n-1)(n-2)(n-3)(n-4)}{120} \right) \\
        &= \frac{s(s-1)(s-2)(10n^{2} -5ns -55n +20s +60)}{n(n-1)(n-2)(n-3)(n-4)} \\
        &= \frac{5s(s-1)(s-2)(2n-s-3)(n-4)}{n(n-1)(n-2)(n-3)(n-4)} \\
        &= \frac{10s(s-1)(s-2)(n-s/2-3/2)}{n(n-1)(n-2)(n-3)}
    \end{align*}
\end{proof}

\subsection{Proof of Lemma~\ref{lem:numerical-distances}}

In order to prove Lemma~\ref{lem:numerical-distances}, we first show the following result.
\begin{lemma}
\label{lem:relevant-recurrences}
Define $\ell_{i, k}$ as in Lemma~\ref{lem:confidence-intervals} with respect to RLS$_k$. Then
$\ell_{i,2} \ge \ell_{i,3}$ as well as $\ell_{i,4} \ge \ell_{i,5}$ and
\begin{align*}
u_{i,1} =\;& u_{i-1,1} - \frac{\ell_{i,1}}{20} + o(n)\\
\ell_{i,3} \ge\;& \ell_{i-1,3} - \frac{3\ell_{i-1, 3}^2}{20n}  - o(n)\\
u_{i,3} \le\;& u_{i-1,3} - \frac{3\ell_{i,3}^2}{20n} + o(n)\\
\ell_{i,5} \ge\;& \ell_{i-1,5} - \frac{10\ell_{i-1, 5}^3}{20n^2}  - o(n).
\end{align*}
\end{lemma}
\begin{proof}
The inequalities $\ell_{i,2} \ge \ell_{i,3}$ and $\ell_{i,4} \ge \ell_{i,5}$ follow from the fact that for even~$k$, $\Delta_k(s) \le \Delta_{k+1}(s)$ for all distances~$s$~\cite[Lemma~27]{paper:opt_param_choices_precise_bba}.

The other results essentially follow from Lemma~\ref{lem:confidence-intervals} along with the drift bounds from Lemma~\ref{lem:drift-bounds}.
The equality for $u_{i,1}$ follows immediately from $\Delta_1(\ell_{i-1,1})=\ell_{i-1,1}/n$.
The lower bound for $\ell_{i, 3}$ follows from $\Delta_3(\ell_{i-1,3}) \le \frac{3\ell_{i-1,3}^2}{n^2}$ and, likewise, the lower bound for $\ell_{i, 5}$ follows from $\Delta_5(\ell_{i-1,5}) \le \frac{10\ell_{i-1,5}^3}{n^3}$.
The upper bound for $u_{i, 3}$ follows from $\Delta_3(\ell_{i,3})  = \frac{3\ell_{i,3}(\ell_{i,3}-1)}{n(n-1)} \ge \frac{3\ell_{i,3}^2}{n^2} - O(1/n)$. Along with a factor of $n/20$, the term $-O(1/n)$ leads to an error term of $-O(1)$ that is absorbed in the $-o(n)$ term.
\end{proof}

\begin{proof}[Proof of Lemma~\ref{lem:numerical-distances}]
We first argue that it is safe to focus on the leading constants in the recurrences given in Lemma~\ref{lem:relevant-recurrences}, that is, that the terms of $o(n)$ can essentially be neglected.
Since the drift $\Delta_k(s)$ is non-decreasing in~$s$, we have $\Delta_k(s - o(n)) \le \Delta_k(s)$ and thus any negative small order terms in the terms $\ell_{i,1}/20$, $3\ell_{i, 3}/(20n)$, and $10\ell_{i, 5}/(20n)$ can be ignored. Every application of a recurrence formula from Lemma~\ref{lem:relevant-recurrences} subtracts another term of $-o(n)$. But since we only consider a constant number of applications, the total error term is still $-o(n)$.

For the upper bounds, it is also not hard to show that $\Delta_k(s + o(n)) \le \Delta_k(s) + o(1)$ for $k \in \{1, 3, 5\}$, which introduces an additional $+o(n)$ term in each application of a recurrence. By the previous arguments, the total error in a constant number of applications sums up to $+o(n)$.

This implies that, modulo small order terms, the distance to the optimum in any period can be bounded by considering the leading constants
$c_{\ell, i, k}$ in $\ell_{i, k}$ and $c_{u, i, k}$ in $u_{i, k}$, when taking the inequalities as equalities. Then $c_{\ell, 0, k} = c_{u, 0, k} = 1/2$ for all $k$ and
$c_{\ell, i, 1} = c_{\ell, i-1, 1} - c_{\ell, i-1, 1}/20,
c_{\ell, i, 3} = c_{\ell, i-1, 3} - 3c_{\ell, i-1, 3}^2/20, c_{\ell, i, 5} = c_{\ell, i-1, 5} - 10c_{\ell, i-1, 5}^3/20$ and $c_{u, i, 3} = c_{u, i-1, 3} - 3c_{\ell, i, 3}^2/20$.

We solved these recurrences numerically by implementing the above formulas in Java. The resulting leading constants were (see Table~\ref{tab:numerical-values} for the complete output; we also show $c_{\ell,80,1}$ and $c_{u, 80,5}$ defined similarly, though we do not need them):
\begin{align*}
[c_{\ell, 80, 1}, c_{u, 80, 1}] =\;& [0.00825768719250682, 0.03284480283288153]\\
[c_{\ell, 80, 3}, c_{u, 80, 3}] =\;& [0.06992905096565742, 0.10669554014031371]\\
[c_{\ell, 80, 5}, c_{u, 80, 5}] =\;& [0.10758784803030164, 0.16946517555735155].
\end{align*}
Noticing that these intervals are non-overlapping, with gaps of order $\Omega(1)$, implies the claim for the stated comparisons of bounds for RLS$_1$, RLS$_3$, and RLS$_5$, even when taking into account error terms of $o(n)$. The results for RLS$_2$ and RLS$_4$ follow immediately from these results along with Lemma~\ref{lem:relevant-recurrences}.

The additional statement about the distance being at most $0.17n$ follows since all $c_{u, 80, k}$ values are less than $0.17 - \Omega(1)$.
%
%
%
\end{proof}

\subsection{Negative Drift Application in the Proof of Theorem~\ref{thm:small_k_wins_long_runs_om}}

In the following we give details omitted from the proof of Theorem~\ref{thm:small_k_wins_long_runs_om}.
\begin{proof}
For cutoff times larger than $4n$, it is possible for the algorithms that lag behind to catch up after time~$4n$.
To this end, we define the distance between two algorithms RLS$_a$, RLS$_b$ with $a < b$ as $D_t^{a,b} := s_{t, b} - s_{t, a}$, where $s_{t, a}$ and $s_{t,b}$ refer to the respective distances to the optimum at time~$t$. Initially we have $D_t^{a,b} = \Omega(n)$ for all considered algorithm pairs. We will apply the negative drift theorem~\cite{Oliveto2011,Oliveto2012Erratum} in the version for self-loops~\cite{Rowe2013} to show that with overwhelming probability $D_t^{a,b}$ does not drop to 0 until RLS$_a$ has found an optimum ($s_{t,a} < a$).


Consider the situation where $D_t^{a,b}$ has decreased to a value at most $n^{1/4}$. We then argue that
\[
\E(D_{t+1}^{a,b} - D_{t}^{a,b} \mid 0 \le D_{t}^{a,b} \le n^{1/4}, s_{t,a} \ge a, s_{t,b}) = \Omega(\Delta_a(s_{t,a})).
\]
For RLS$_1$ and RLS$_3$ the above expectation is at least (using Lemma~\ref{lem:drift-bounds} and $s_{t, 1} \le 0.17n$)
\begin{align*}
& \Delta_1(s_{t,1}) - \Delta_3(s_{t,3})
\ge \frac{s_{t,1}}{n} - \frac{3(s_{t,1}+n^{1/4})^2}{n^2}\\
=\;&\frac{s_{t,1}}{n} \left(1 - \frac{3s_{t,1}}{n} - o(1)\right)
\ge\frac{s_{t,1}}{n} \left(1 - 3 \cdot 0.17 - o(1)\right)
= \Omega(\Delta_1(s_{t,1})).
\end{align*}
For RLS$_3$ and RLS$_5$ the above expectation is at least (using Lemma~\ref{lem:drift-bounds} and $s_{t, 3} \le 0.17n$)
\begin{align*}
\Delta_3(s_{t,3}) - \Delta_5(s_{t,5})
\ge\;& \frac{3s_{t,3}(s_{t,3}-1)}{n(n-1)} - \frac{10(s_{t,3}+n^{1/4})^3}{n^3}\\
=\;& \frac{3s_{t,3}(s_{t,3}-1)}{n(n-1)} - \frac{3s_{t,3}^2}{n^2} \left(\frac{10s_{t,3}}{3n} + o(1)\right)\\
=\;& \Omega(\Delta_3(s_{t,3}))
\end{align*}
The statement also follows for even~$b$ as $\Delta_b(s) < \Delta_{b+1}(s)$.

We also have $\Delta_k(s)/k \le \prob(s_{t+1,k} < s_{t,k}) \le \Delta_k(s)$ for all $k,s$. The above calculations have further established $\Delta_b(s_{t,b}) = O(\Delta_a(s_{t,a}))$. Hence $\prob(D_{t+1}^{a,b} \neq D_{t}^{a,b}) = \Theta(\Delta_a(s_{t,a}))$.

Together, this implies that the first condition of the negative drift theorem with self-loops~\cite{Rowe2013} is satisfied with respect to $D_t^{a,b}$ and the interval $[0, n^{1/4}]$. The second condition is trivial as the jump length is bounded by~$b = O(1)$. Applying said theorem yields that probability of RLS$_b$ catching up to RLS$_a$ before RLS$_a$ finds an optimum in $2^{\Omega(n^{1/4})}$ generations is $e^{-\Omega(n^{1/4})}$. By Markov's inequality, the probability that RLS$_a$ has not found an optimum within this time is $e^{-\Omega(n^{1/4})}$ as well. Summing up all failure probabilities proves the claim.
\end{proof}

\subsection{Comparison of RLS\boldmath$_2$ and RLS$_3$ in Lemma~\ref{lem:RLS3_beats_RLS1_short_race_wop}}

\begin{proof}[Proof of Lemma~\ref{lem:RLS3_beats_RLS1_short_race_wop} for RLS$_2$ and RLS$_3$]
Let $s_{t,2}$ be the distance to the optimum in RLS$_2$ and $s_{t,3}$ be the distance to the optimum in RLS$_3$ at time~$t$.
Let $\varepsilon > 0$ be a constant chosen later, then by Chernoff bounds,
    \[ \prob(s_{0,2}, s_{0,3} \in [(n-\varepsilon\kappa) / 2, (n+\varepsilon\kappa) / 2]) \ge 1 - 4e^{-\Omega(\kappa^2/n)} \]
We assume in the following that this is the case.
Then RLS$_3$ wins if in $\kappa$ steps RLS$_3$'s progress exceeds that of
RLS$_2$ by at least $\varepsilon \kappa$.


Define $D_t := (s_{t+1,3}-s_{t,3})-(s_{t+1,2}-s_{t,2})$ to be the difference in the progress values made by the two algorithms.
Note that
\begin{align*}
    \E(D_t) =\;& \frac{s_{t,3}}{n} \cdot \frac{3(s_{t,3}-1)}{n-1} - \frac{2 s_{t,2}(s_{t,2} - 1)}{n(n-1)}\\
    =\;& 3 \left(\frac{s_{t,3}}{n}\right)^2 - 2 \left( \frac{s_{t,2} - 1}{n-1} \right)^{2} - O(1/n).
\end{align*}
Note that the leading constant in $\kappa$ is chosen as $0.03 < \gamma := 1/3 \cdot (1/2-1/\sqrt{6})$. This implies that for $t \le 0.03n$ we always have $s_{t,2} \le n/2 + \varepsilon \kappa \le n/2 + \varepsilon n$, and therefore $s_{t,2} - 1 \le n/2 + \varepsilon n - 1$ and
    \begin{align*}
        \frac{s_{t,2} - 1}{n-1} &\le \frac{n/2 + \varepsilon n - 1}{n-1} = \frac{n + 2\varepsilon n - 2}{2(n-1)} = \frac{(n-1)+(2\varepsilon n - 1)}{2(n-1)} \\
        &= \frac{1}{2} + \frac{2 \varepsilon n - 1}{2(n-1)} \le \frac{1}{2} + \frac{\varepsilon n}{n-1} \le \frac{1}{2} + 2 \varepsilon
    \end{align*}
    for $n \ge 2$.
We also have $s_{t,3} \ge n/2-\varepsilon \kappa -0.09n$.
We bound the latter using $\varepsilon \kappa \le \varepsilon n$ and $0.09n = 3\gamma n - 3(\gamma-0.03)n \le 3\gamma n - 4\varepsilon n$ if we choose $\varepsilon$ small enough:
\[
    s_{t,3} \ge n/2 - \varepsilon n - (1/2-1/\sqrt{6})n + 4\varepsilon n =  n/\sqrt{6} + 3\varepsilon n.
\]
Using these inequalities,
\begin{align*}
    \E(D_t) \ge\;& 3(1/\sqrt{6} + 3\varepsilon)^2 - 2(1/2+2\varepsilon)^{2} - O(1/n)\\
    \ge\;& (3\sqrt{6} - 4)\varepsilon + 19\varepsilon^{2} - O(1/n)
\end{align*}
Now, for $D := \sum_{t=1}^\kappa D_t$, using $\E(D)  \ge \varepsilon \kappa + (3\sqrt{6} - 5)\varepsilon \kappa + 19\varepsilon^{2}\kappa - O(\kappa/n) = \varepsilon \kappa + \Omega(\kappa)$,
\[
\prob(D \le \varepsilon \kappa) \le
\prob(D \le \E(D) - \Omega(\kappa)).
\]
By the method of bounded differences~\cite[Theorem 3.67]{paper:scheideler_hab_thesis}, this is at most
$    \exp(-\Omega(\kappa^2)/\Theta(\kappa))
    = \exp(-\Omega(\kappa))$.

\end{proof}


\begin{table*}
\centering
\[
\footnotesize
\begin{array}{cccc}
i & [c_{\ell, i, 1}, c_{u, i, 1}] & [c_{\ell, i, 3}, c_{u, i, 3}] & [c_{\ell, i, 5}, c_{u, i, 5}]\\
\toprule
0 & [0.5, 0.5] & [0.5, 0.5] & [0.5, 0.5]\\
1 & [0.475, 0.47625] & [0.4625, 0.4679140625] & [0.4375, 0.4581298828125]\\
2 & [0.45125, 0.4536875] & [0.4304140625, 0.4401256227203369] & [0.3956298828125, 0.427167293913044]\\
3 & [0.4286875, 0.432253125] & [0.4026256227203369, 0.415809513909696] & [0.364667293913044, 0.4029201579795064]\\
4 & [0.407253125, 0.41189046875] & [0.378309513909696, 0.3943418006625074] & [0.3404201579795064, 0.38319521251708305]\\
5 & [0.38689046875, 0.39254594531250003] & [0.3568418006625074, 0.3752413900574983] & [0.32069521251708305, 0.3667041957126408]\\
6 & [0.3675459453125, 0.37416864804687505] & [0.3377413900574983, 0.35813100307380263] & [0.3042041957126408, 0.3526286382685892]\\
7 & [0.34916864804687503, 0.3567102156445313] & [0.32063100307380266, 0.3427103670539857] & [0.2901286382685892, 0.3404179033516475]\\
8 & [0.3317102156445313, 0.3401247048623047] & [0.30521036705398574, 0.3287373618304014] & [0.2779179033516475, 0.3296849416774596]\\
9 & [0.3151247048623047, 0.32436846961918947] & [0.29123736183040144, 0.3160144816915116] & [0.2671849416774596, 0.3201480700149588]\\
10 & [0.29936846961918945, 0.30940004613823] & [0.2785144816915116, 0.3043789342147279] & [0.2576480700149588, 0.3115964049062649]\\
11 & [0.28440004613823, 0.2951800438313185] & [0.2668789342147279, 0.29369527938558954] & [0.24909640490626486, 0.30386831113365304]\\
12 & [0.2701800438313185, 0.2816710416397526] & [0.25619527938558956, 0.28384987620867047] & [0.241368311133653, 0.29683741375164874]\\
13 & [0.2566710416397526, 0.268837489557765] & [0.24634987620867052, 0.2747466369824664] & [0.23433741375164874, 0.2904032086307019]\\
14 & [0.24383748955776494, 0.2566456150798767] & [0.23724663698246642, 0.26630374196854284] & [0.2279032086307019, 0.28448457683091705]\\
15 & [0.2316456150798767, 0.24506333432588287] & [0.2288037419685429, 0.2584510691177217] & [0.22198457683091705, 0.2790151929249044]\\
16 & [0.22006333432588285, 0.23406016760958873] & [0.22095106911772175, 0.2511281628760821] & [0.21651519292490443, 0.2739402035975807]\\
17 & [0.2090601676095887, 0.2236071592291093] & [0.2136281628760821, 0.2442826140800106] & [0.21144020359758073, 0.2692137792672586]\\
18 & [0.19860715922910926, 0.21367680126765382] & [0.2067826140800106, 0.23786875665714619] & [0.20671377926725862, 0.2647972787503718]\\
19 & [0.1886768012676538, 0.20424296120427113] & [0.2003687566571462, 0.23184661086049657] & [0.20229727875037182, 0.26065785271652]\\
20 & [0.1792429612042711, 0.19528081314405757] & [0.1943466108604966, 0.22618102008755236] & [0.19815785271652, 0.25676736662721406]\\
21 & [0.17028081314405755, 0.1867667724868547] & [0.1886810200875524, 0.22084094098636045] & [0.19426736662721408, 0.2531015598999752]\\
22 & [0.16176677248685467, 0.17867843386251195] & [0.18334094098636047, 0.21579885589009584] & [0.1906015598999752, 0.2496393821883677]\\
23 & [0.15367843386251193, 0.17099451216938635] & [0.17829885589009586, 0.21103028358833825] & [0.18713938218836773, 0.2463624641540121]\\
24 & [0.14599451216938633, 0.16369478656091704] & [0.17353028358833827, 0.2065133696900009] & [0.18386246415401214, 0.2432546915548695]\\
25 & [0.13869478656091702, 0.15676004723287118] & [0.16901336969000091, 0.20222854181990554] & [0.18075469155486953, 0.24030185954935943]\\
26 & [0.13176004723287116, 0.15017204487122762] & [0.16472854181990557, 0.1981582179463887] & [0.17780185954935945, 0.23749138989398264]\\
27 & [0.1251720448712276, 0.14391344262766625] & [0.1606582179463887, 0.19428655849733228] & [0.17499138989398266, 0.23481209790126797]\\
28 & [0.11891344262766622, 0.13796777049628295] & [0.1567865584973323, 0.19059925475851666] & [0.172312097901268, 0.23225399909908556]\\
29 & [0.11296777049628291, 0.1323193819714688] & [0.15309925475851668, 0.18708334748737468] & [0.16975399909908556, 0.22980814781389072]\\
30 & [0.10731938197146877, 0.12695341287289535] & [0.1495833474873747, 0.18372707081054537] & [0.16730814781389072, 0.22746650161144188]\\
31 & [0.10195341287289533, 0.1218557422292506] & [0.1462270708105454, 0.18051971737487055] & [0.16496650161144188, 0.22522180682417325]\\
32 & [0.09685574222925057, 0.11701295511778806] & [0.14301971737487057, 0.17745152144117238] & [0.16272180682417325, 0.22306750138446313]\\
33 & [0.09201295511778804, 0.11241230736189865] & [0.1399515214411724, 0.17451355718811754] & [0.16056750138446313, 0.2209976319460385]\\
34 & [0.08741230736189863, 0.10804169199380372] & [0.13701355718811756, 0.1716976499601163] & [0.1584976319460385, 0.21900678286847072]\\
35 & [0.08304169199380369, 0.10388960739411353] & [0.13419764996011632, 0.1689962985718936] & [0.15650678286847072, 0.2170900151036049]\\
36 & [0.0788896073941135, 0.09994512702440786] & [0.13149629857189363, 0.16640260709117732] & [0.1545900151036049, 0.21524281338839488]\\
37 & [0.07494512702440784, 0.09619787067318747] & [0.12890260709117735, 0.16391022477394196] & [0.15274281338839488, 0.21346104043872052]\\
38 & [0.07119787067318745, 0.0926379771395281] & [0.12641022477394198, 0.16151329303483217] & [0.15096104043872052, 0.21174089707039276]\\
39 & [0.06763797713952807, 0.08925607828255169] & [0.1240132930348322, 0.15920639850743068] & [0.14924089707039276, 0.2100788873595957]\\
40 & [0.06425607828255167, 0.08604327436842411] & [0.12170639850743073, 0.15698453139178326] & [0.1475788873595957, 0.208471788105288]\\
41 & [0.061043274368424084, 0.0829911106500029] & [0.1194845313917833, 0.15484304840549615] & [0.145971788105288, 0.20691662197811622]\\
42 & [0.05799111065000288, 0.08009155511750277] & [0.1173430484054962, 0.15277763975413194] & [0.14441662197811622, 0.20541063383999505]\\
43 & [0.05509155511750274, 0.07733697736162763] & [0.11527763975413201, 0.15078429962003942] & [0.14291063383999505, 0.2039512698001972]\\
44 & [0.0523369773616276, 0.07472012849354626] & [0.1132842996200395, 0.148859299738979] & [0.1414512698001972, 0.20253615864110383]\\
45 & [0.04972012849354622, 0.07223412206886895] & [0.11135929973897907, 0.14699916569322563] & [0.14003615864110383, 0.20116309530246781]\\
46 & [0.04723412206886891, 0.06987241596542551] & [0.1094991656932257, 0.14520065560009876] & [0.13866309530246781, 0.19983002615933385]\\
47 & [0.04487241596542546, 0.06762879516715424] & [0.10770065560009882, 0.1434607409175951] & [0.13733002615933385, 0.19853503586738613]\\
48 & [0.04262879516715419, 0.06549735540879653] & [0.10596074091759516, 0.14177658912522423] & [0.13603503586738613, 0.19727633558184743]\\
49 & [0.04049735540879648, 0.06347248763835671] & [0.1042765891252243, 0.1401455480692856] & [0.13477633558184743, 0.19605225238325436]\\
50 & [0.03847248763835666, 0.06154886325643888] & [0.10264554806928568, 0.138565131788519] & [0.13355225238325436, 0.19486121976638196]\\
51 & [0.03654886325643883, 0.05972142009361694] & [0.10106513178851907, 0.1370330076590044] & [0.13236121976638196, 0.1937017690680174]\\
52 & [0.03472142009361689, 0.05798534908893609] & [0.09953300765900447, 0.13554698471695728] & [0.1312017690680174, 0.19257252172578224]\\
53 & [0.03298534908893604, 0.05633608163448929] & [0.09804698471695736, 0.1341050030351442] & [0.13007252172578224, 0.1914721822742598]\\
54 & [0.03133608163448924, 0.054769277552764825] & [0.09660500303514427, 0.13270512404343116] & [0.1289721822742598, 0.19039953199669632]\\
55 & [0.02976927755276478, 0.05328081367512658] & [0.09520512404343123, 0.13134552169681238] & [0.12789953199669632, 0.1893534231608392]\\
56 & [0.02828081367512654, 0.051866772991370255] & [0.09384552169681247, 0.13002447440543036] & [0.1268534231608392, 0.18833277377632338]\\
57 & [0.026866772991370212, 0.050523434341801746] & [0.09252447440543043, 0.1287403576508302] & [0.12583277377632338, 0.18733656281864094]\\
58 & [0.025523434341801703, 0.04924726262471166] & [0.09124035765083026, 0.12749163722119247] & [0.12483656281864094, 0.18636382587131584]\\
59 & [0.024247262624711618, 0.04803489949347608] & [0.08999163722119255, 0.12627686300572985] & [0.12386382587131585, 0.18541365114361122]\\
60 & [0.023034899493476035, 0.046883154518802275] & [0.08877686300572994, 0.12509466329495914] & [0.12291365114361123, 0.18448517582605053]\\
61 & [0.021883154518802232, 0.04578899679286216] & [0.08759466329495923, 0.12394373953929555] & [0.12198517582605055, 0.1835775827503444]\\
62 & [0.02078899679286212, 0.04474954695321906] & [0.08644373953929564, 0.12282286152346492] & [0.1210775827503444, 0.18269009732407562]\\
63 & [0.019749546953219014, 0.04376206960555811] & [0.08532286152346501, 0.12173086291868206] & [0.12019009732407564, 0.18182198471378214]\\
64 & [0.018762069605558065, 0.04282396612528021] & [0.08423086291868215, 0.12066663717847818] & [0.11932198471378216, 0.18097254725295675]\\
65 & [0.01782396612528016, 0.0419327678190162] & [0.08316663717847828, 0.1196291337475417] & [0.11847254725295678, 0.18014112205401098]\\
66 & [0.016932767819016155, 0.04108612942806539] & [0.08212913374754177, 0.11861735455602346] & [0.11764112205401099, 0.17932707880547316]\\
67 & [0.016086129428065348, 0.04028182295666212] & [0.08111735455602354, 0.11763035077449831] & [0.11682707880547319, 0.178529817737651]\\
68 & [0.01528182295666208, 0.03951773180882902] & [0.08013035077449839, 0.11666721980721169] & [0.11602981773765104, 0.17774876774171805]\\
69 & [0.014517731808828975, 0.03879184521838757] & [0.07916721980721177, 0.11572710250341119] & [0.11524876774171808, 0.17698338462871452]\\
70 & [0.013791845218387526, 0.03810225295746819] & [0.07822710250341126, 0.1148091805684993] & [0.11448338462871455, 0.17623314951630878]\\
71 & [0.01310225295746815, 0.037447140309594784] & [0.07730918056849938, 0.11391267415847338] & [0.1137331495163088, 0.17549756733236907]\\
72 & [0.012447140309594743, 0.036824783294115045] & [0.07641267415847346, 0.11303683964266602] & [0.1129975673323691, 0.17477616542546548]\\
73 & [0.011824783294115005, 0.036233544129409295] & [0.0755368396426661, 0.11218096752118574] & [0.1122761654254655, 0.17406849227337523]\\
74 & [0.011233544129409256, 0.03567186692293883] & [0.07468096752118582, 0.11134438048470068] & [0.11156849227337524, 0.17337411628151428]\\
75 & [0.010671866922938793, 0.0351382735767919] & [0.07384438048470075, 0.11052643160532528] & [0.1108741162815143, 0.17269262466397758]\\
76 & [0.010138273576791854, 0.0346313598979523] & [0.07302643160532536, 0.10972650264837419] & [0.11019262466397761, 0.1720236224005497]\\
77 & [0.00963135989795226, 0.03414979190305469] & [0.07222650264837427, 0.10894400249565185] & [0.10952362240054973, 0.1713667312636564]\\
78 & [0.009149791903054648, 0.03369230230790196] & [0.07144400249565193, 0.10817836567176205] & [0.10886673126365642, 0.17072158890977368]\\
79 & [0.008692302307901915, 0.033257687192506866] & [0.07067836567176213, 0.10742905096565734] & [0.10822158890977371, 0.1700878480303016]\\
80 & [0.00825768719250682, 0.03284480283288153] & [0.06992905096565742, 0.10669554014031371] & [0.10758784803030164, 0.16946517555735155]\\
\bottomrule
\end{array}
\]
\caption{Numerical values for leading constants in progress bounds from Lemma~\ref{lem:numerical-distances}.}
\label{tab:numerical-values}
\end{table*}